# Fast and Scalable Learning of Sparse Changes in High-Dimensional Gaussian Graphical Model Structure

**Beilun Wang**     **Arshdeep Sekhon**     **Yanjun Qi**

Department of Computer Science, University of Virginia

www.jointggm.org

## Abstract

We focus on the problem of estimating the change in the dependency structures of two $p$-dimensional Gaussian Graphical models (GGMs). Previous studies for sparse change estimation in GGMs involve expensive and difficult non-smooth optimization. We propose a novel method, DIFFEE for estimating <u>DIFF</u>erential networks via an <u>E</u>lementary <u>E</u>stimator under a high-dimensional situation. DIFFEE is solved through a faster and closed form solution that enables it to work in large-scale settings. We conduct a rigorous statistical analysis showing that surprisingly DIFFEE achieves the same asymptotic convergence rates as the state-of-the-art estimators that are much more difficult to compute. Our experimental results on multiple synthetic datasets and one real-world data about brain connectivity show strong performance improvements over baselines, as well as significant computational benefits.

## 1 Introduction

Learning the change of interactions between random variables is an essential task in many real-world applications. For instance, identifying the difference in brain connectivity networks of subjects from different groups can shed light on understanding psychiatric diseases [10]. As another example in gene expression analysis, interests may not center on a particular graph representing interactions among genes, but instead on how gene interactions change when external stimuli change [14]. Such change detection can significantly simplify network-driven studies about diseases, drugs or system understanding.

In this paper we consider Gaussian graphical models (GGMs) and focus on estimating changes in the dependency structure of two $p$-dimensional GGMs, based on $n_c$ and $n_d$ samples drawn from the models, respectively. Recent literature has made significant advances on estimating the statistical dependency structure of GGMs based on samples drawn from the model [1][13] (reviewed in 2.1). Detecting structural changes naturally involves two sets of data samples. Given two sets of data (in the form of two matrices) $\mathbf{X}_c \in \mathbb{R}^{n_c \times p}$ and $\mathbf{X}_d \in \mathbb{R}^{n_d \times p}$ identically and independently drawn from normal distributions $N_p(\mu_c, \Sigma_c)$ and $N_p(\mu_d, \Sigma_d)$ respectively, our goal is to estimate the structural change $\Delta$ (defined by [35]) [1]:

$$\Delta = \Omega_d - \Omega_c \quad (1.1)$$

Here $\mu_c, \mu_d \in \mathbb{R}^p$ describes the mean and $\Sigma_c, \Sigma_d \in \mathbb{R}^{p \times p}$ represents covariance matrices. In Eq. (1.1), the precision matrix $\Omega_c := (\Sigma_c)^{-1}$ and $\Omega_d := (\Sigma_d)^{-1}$. The conditional dependency graph structure of a GGM is encoded by the sparsity pattern of its precision matrix. The entries of $\Delta$ describe if the magnitude of conditional dependency of a pair of random variables changes between two conditions. They can also be interpreted as the differences in the partial covariance of each pair of random variables between the two conditions.

In particular, we focus on estimating the change $\Delta$ under a high-dimensional situation, where the number of variables $p$ may exceed the number of observations: $p > \max(n_c, n_d)$. In such high-dimensional settings, it is still possible to conduct consistent estimation by leveraging low-dimensional structure such as sparsity constraints. A sparse $\Delta$ indicates few of its entries are non-zero. In the context of estimating structural changes of two GGMs, this translates into a differential network with few edges. However, we do not assume the individual structures $\Omega_c$ and $\Omega_d$ to be sparse, and they may both correspond to dense matrices. Our main objective is to get an estimated $\widehat{\Delta}$ of the true change $\Delta^*$ such that the estimation error $(\widehat{\Delta} - \Delta^*)$ is bounded.

---

[1]Using which of the two sample sets as 'c' set (or 'd' set) does not affect the computational cost and the statistical convergence rates of our model. For instance, on samples from a controlled disease study 'c' may represent the 'control' group and 'd' may represent the 'disease' group.

A naive approach to detecting structural changes in GGMs is a two-step procedure in which we estimate $\widehat{\Omega}_d$ and $\widehat{\Omega}_c$ from two sets of samples separately and obtain $\widehat{\Delta} = \widehat{\Omega}_d - \widehat{\Omega}_c$. However, in a high-dimensional setting, this strategy needs to assume that both $\Omega_d$ and $\Omega_c$ are sparse (in order to achieve consistent estimation). This is not necessarily true even if the change $\Delta$ is sparse. A motivating example from identifying the difference in connectivity networks among brain regions (functional networks) of subjects from different groups. Recent literature in neuroscience has suggested functional networks are not sparse. On the other hand, differences in functional connections across subjects should be sparse [2]. In the application of estimating genetic networks of two conditions, each individual network might contain hub nodes and therefore not entirely sparse.

This has motivated a few recent studies to directly estimate the changes of structures from two sets of samples. Zhang et al. used the fused norm for regularizing maximum likelihood estimation (MLE) to simultaneously learn two GGMs with a sparsity-inducing penalty on the difference [33]. The resulting penalized MLE framework is a log-determinant program, which can be solved by block coordinate descent algorithms [33] or the alternating direction method of multipliers (ADMM) by the JGL-fused package [9]. Later Liu et al. proposed to use density ratio estimation (DRE) to directly learn structural changes without having to identify the structures of each individual graphical model. The authors focused on exponential family-based pairwise Markov networks [17] and solve the resulting optimization using proximal gradient descent [16]. A follow-up study showed that under certain conditions the DRE method recovers the correct parameter sparsity with high probability [16]. More recently, Fazayeli et al. introduced a regularized density ratio estimator for direct structured change estimation in Ising model structure. Theoretically, the authors showed that the estimation error converges to zero under milder conditions than DRE [12]. Another related regularized convex program to directly learn structural changes without going through the learning of the individual GGMs is the Diff-CLIME method [34]. Diff-CLIME uses an $\ell_1$ minimization formulation constrained by the covariance-precision matching. Diff-CLIME reduces the estimation problem to solving linear programs (LP) and can be solved by using any standard LP solvers. Another recent work relaxes the Gaussian assumption in Diff-CLIME model to a semiparametric distribution [30]. All previous studies have used $\ell_1$ regularized convex formulation for estimating structural changes. While state-of-the-art optimization methods have been developed to solve the resulting non-smooth programs, their iterative algorithms are very expensive for large-scale problems.

In this paper, we propose a simple estimator, namely <u>DIFF</u>erential networks via an <u>E</u>lementary <u>E</u>stimator (DIFFEE) for fast and scalable learning of sparse structural change in high-dimensional GGMs. Briefly speaking, DIFFEE provides the following benefits:

- **Novel approach:** DIFFEE presents a novel way of structural change estimation by extending the elementary estimator for sparse GGM[31]. (Section 2.3)
- **Closed-Form optimization:** We optimize DIFFEE through a closed-form manner that can dramatically improve its entire time complexity to $O(p^3)$. The closed-form solution makes DIFFEE scalable to much larger values of $p$, compared to the aforementioned state-of-the-art. (Section 2.4)
- **Convergence rate:** We theoretically prove that DIFFEE achieves the same sharp convergence rate as the aforementioned regularized convex programs. (Section 2.5)
- **Evaluation:** DIFFEE is evaluated using several simulated datasets and one real-world neuroscience dataset. It improves the state-of-the-art baselines with better estimation F-1 scores as well as significant computational advantages. (Section 3)

## 2 Method

Our main goal is to design a simple estimator with closed-form solutions that can achieve the same sharp convergence rates as the state-of-the-art regularized convex formulations under high-dimensional settings. This has been achieved by the so-called elementary estimators [31] in the context of learning sparse GGM from one set of samples. Inspired by [31], naturally, we ask the following question: *is there an elementary estimator that can estimate structural changes in GGMs with a closed-form solution and achieves the near-optimal convergence rate?* This section provides an affirmative response of "YES" by proposing the DIFFEE algorithm. In the rest of this section, we first review the background of elementary estimator and then propose to estimate the differential network through an elementary estimator. Finally, we provide a rigorous theoretical analysis of DIFFEE's convergence rates.

*Notations:* Given a $p$-dimensional vector $x = (x_1, x_2, \ldots, x_p)^T \in \mathbb{R}^p$, $\| x \|_1 = \sum_i |x_i|$ represents the $\ell_1$-norm of $x$. $\| x \|_\infty = \max_i |x_i|$ is the $\ell_\infty$-norm of $x$. $\| x \|_2 = \sqrt{\sum_i x_i^2}$ describes the $\ell_2$-norm of $x$.

## 2.1 Background: Elementary Estimator for Estimating Sparse GGM in Closed Form

Sparse Gaussian Graphical Model(sGGM)[15, 18, 32] assumes data samples are independently and identically drawn from $N_p(\mu, \Sigma)$, a multivariate normal distribution with mean $\mu$ and covariance matrix $\Sigma$. The conditional dependency graph structure among its $p$ random variables is encoded by the sparsity pattern of the inverse covariance matrix (precision matrix) $\Omega$. $\Omega := (\Sigma)^{-1}$. An edge does not connect $j$-th node (variable) and $k$-th node (variable) if and only if $\Omega_{jk} = 0$ (i.e., conditionally independent). sGGM imposes an $\ell_1$ penalty on the parameter $\Omega$.

**Regularized MLE:** Over the past decade, significant progress has been made on estimating sGGMs based on samples drawn from the model. Most sGGM estimation [1, 32] are based on minimizing the $\ell_1$-regularized Gaussian negative log likelihood:

$$\underset{\Omega}{\operatorname{argmin}} - \log \det(\Omega) + <\Omega, \Sigma> + \lambda_n ||\Omega||_1 \quad (2.1)$$

Friedman et al. [13] used a blockwise coordinate descent algorithm called the graphical lasso to efficiently solve the regularized MLE formulation. Alternatively, Meinshausen et al. [19] introduced a neighborhood selection approach that applies a lasso linear regression on each variable separately and combines the result to learn the conditional dependency structure.

**CLIME:** Later Cai et al. [5] proposed a constrained $\ell_1$ minimization method for inverse matrix estimation (abbreviated as CLIME) formulated as follows:

$$\underset{\Omega}{\operatorname{argmin}} ||\Omega||_1$$
$$\text{subject to: } ||\Sigma \Omega - I||_\infty \leq \lambda_n \quad (2.2)$$

The above formulation can be decomposed into column-wise linear programming. However the computational cost of this LP formulation gets significantly demanding as $p$ increases.

**EE-sGGM:** Recently, Yang et al. [31] proposed a closed-form estimator for learning Gaussian graphical models through the following form:

$$\underset{\Omega}{\operatorname{argmin}} ||\Omega||_{1,,\text{off}}$$
$$\text{subject to:} ||\Omega - [T_v(\widehat{\Sigma})]^{-1}||_{\infty,\text{off}} \leq \lambda_n \quad (2.3)$$

Eq. (2.3) is a special case of the elementary estimator for graphical models (GM) of exponential families proposed in [31], namely Elementary Estimators-GM. It has the following generic formulation:

$$\underset{\theta}{\operatorname{argmin}} ||\theta||_1$$
$$\text{Subject to: } ||\theta - \mathcal{B}^*(\widehat{\phi})||_\infty \leq \lambda_n \quad (2.4)$$

Here $\mathcal{B}^*(\widehat{\phi})$ is the so-called proxy of backward mapping for the target GM (more details in Section A.1). $\lambda_n$ is a regularization parameter. $\widehat{\phi}$ is the empirical mean of the sufficient statistics. For example, in the case of

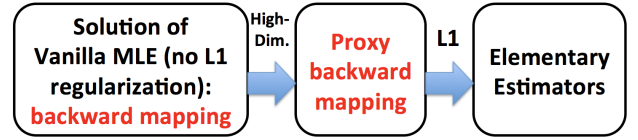

Figure 1: Basic idea of elementary estimators for graphical model.

Gaussian GM, $\widehat{\phi}$ is the sample covariance matrix.

The key idea in Eq. (2.4) (summarized in Figure 1) is to investigate the vanilla MLE and where it "breaks down" for estimating a graphical model of exponential families in the case of high-dimensions [31]. Essentially the vanilla graphical model MLE can be expressed as a backward mapping that computes the model parameters from some given moments in an exponential family distribution. For instance, in the case of learning GGM with vanilla MLE, the backward mapping is $\widehat{\Sigma}^{-1}$ that estimates $\Omega$ from the sample covariance matrix (moment) $\widehat{\Sigma}$.

However, this backward mapping is not available in a closed form for many classes of graphical models, such as Ising model. Even if it has a simple closed form, the backward mapping is normally not well-defined in high-dimensional settings. In the case of GGM, when given the sample covariance $\widehat{\Sigma}$, we cannot just compute the vanilla MLE solution as $[\widehat{\Sigma}]^{-1}$ since $\widehat{\Sigma}$ is rank-deficient when $p > n$. Therefore Yang et al. [31] used carefully constructed proxy backward maps for Eq. (2.4) that are both available in closed-form, and well-defined in high-dimensional settings for GGM and Ising models. $[T_v(\widehat{\Sigma})]^{-1}$ in Eq. (2.3) is the proxy backward mapping Yang et al. used for GGM (more details in Section 2.3 and in Appendix Section A.1).

When given the term $\mathcal{B}^*(\widehat{\phi})$ in Eq. (2.4), the solution of Eq. (2.4) is closed-form and involves only simple thresholding operations. This solution is

$$\widehat{\theta} = S_{\lambda_n}(\mathcal{B}^*(\widehat{\phi}))$$

where the function $S(\cdot)$ is an element-wise soft-thresholding with parameter $\lambda$:

$$[S_\lambda(A)]_{ij} = \text{sign}(A_{ij}) \max(|A_{ij}| - \lambda, 0) \quad (2.5)$$

The optimization in Eq. (2.4) is decomposable into independent element-wise subproblems. Each subproblem corresponds to soft-thresholding. Essentially the final estimators are obtained by performing simple thresholding operations on the proxy backward maps. This class of estimators is thus both computationally practical and highly scalable. Using the theoretical framework proposed by [20] for regularized M-estimators, Yang et al. further proved that the resulting algorithms achieve strong statistical guarantees with sharp convergence rates.

## 2.2 Previous Estimators for Change Estimation in GGM Structure

Multiple estimators have been proposed to estimate sparse differential network from two sets of samples.

**FusedGLasso (Regularized MLE):** The most straightforward estimator for differential network was to extend the classic Graphical lasso estimator [32] for sparse GGM with an added sparsity penalty on the differential network (i.e., fused norm).

$$\operatorname*{argmin}_{\Omega_c, \Omega_d \succ 0, \Delta} n_c(-\log \det(\Omega_c) + <\Omega_c, \widehat{\Sigma}_c>) \\ + n_d(-\log \det(\Omega_d) + <\Omega_d, \widehat{\Sigma}_d>) \\ + \lambda_2(||\Omega_c||_1 + ||\Omega_d||_1) + \lambda_n||\Delta||_1 \quad (2.6)$$

This was solved by block coordinate descent algorithms in [33]. Later the alternating direction method of multipliers (ADMM) was used to solve Eq. (2.6) that needs to run SVD in one sub-procedure [9].

**Diff-CLIME:** Another recent study [35] extended the CLIME estimator to directly learn the $\Delta$ through a constrained optimization formulation.

$$\operatorname*{argmin}_{\Delta} ||\Delta||_1 \\ \text{Subject to: } ||\widehat{\Sigma}_c \Delta \widehat{\Sigma}_d - (\widehat{\Sigma}_c - \widehat{\Sigma}_d)||_\infty \leq \lambda_n \quad (2.7)$$

This reduces the estimation to solving multiple linear programming problems.

**DensityRatio:** The third category of estimators optimizes the following loss:

$$\operatorname*{argmax}_{\Delta} \mathcal{L}_{\text{KLIEP}}(\Delta) - \lambda_n \parallel \Delta \parallel_1 - \lambda_2 \parallel \Delta \parallel_2 \quad (2.8)$$

Here KLIEP minimizes the KL divergence between the true probability density $p_d(x)$ and the estimated $\widehat{p}_d(x) = r(x; \Delta) p_c(x)$ without explicitly modeling the true $p_c(x)$ and $p_d(x)$. Its key idea is the formulation of density ratio term $r(x; \Delta)$ for directly estimating sparse differential network of graphical models in exponential families. This DensityRatio estimator uses the elastic-net penalty for enforcing $\Delta$ to be sparse. The resulting optimization was solved using proximal gradient descent methods in [16].

## 2.3 Proposed Method: DIFFEE

The aforementioned studies cannot avoid certain steps involving expensive computation in their iterative optimization, such as SVD operations in the FusedGLasso, linear programming in the Diff-CLIME, and calculating the normalization term in the Density-Ratio estimator. We aim to propose a scalable and theoretically-guaranteed estimator for estimating sparse differential network under large-scale settings.

**Differential Network by Elementary Estimators (DIFFEE):** Computationally elementary estimators are much faster than their regularized convex program peers for graphical model estimation. Therefore we extend it to the following general estimator for estimating sparse change in GGM structure:

$$\operatorname*{argmin}_{\Delta} ||\Delta||_1 \\ \text{Subject to: } ||\Delta - \mathcal{B}^*(\widehat{\Sigma}_d, \widehat{\Sigma}_c)||_\infty \leq \lambda_n \quad (2.9)$$

The basic idea in Eq. (2.9) is to use a well-defined proxy function $\mathcal{B}^*(\widehat{\Sigma}_d, \widehat{\Sigma}_c)$ to approximate the backward mapping (the vanilla graphical model MLE solution), so that $\mathcal{B}^*(\widehat{\Sigma}_d, \widehat{\Sigma}_c)$ is both well-defined under high-dimensional situations and also has a simple closed-form.

As shown by Figure 1, there are three components in the estimation pipeline of elementary estimator for GM: (1) Backward mapping that is the vanilla MLE solution for estimating an exponential graphical model; (2) Proxy backward mapping $\mathcal{B}^*(\widehat{\Sigma}_d, \widehat{\Sigma}_c$ for dimensional settings; and (3) The closed-form solution of Eq. (2.9) as the final estimator.

**(1) Backward Mapping:** The density ratio of two Gaussian distributions is naturally an exponential-family distribution (see Section A.1.1). Based on [29], learning an exponential family distribution from data means to estimate its canonical parameter. For an exponential family distribution, computing the canonical parameter through vanilla graphical model MLE can be expressed as a backward mapping (the first step in Figure 1). Through simple derivations in Eq. (A.8), we can easily conclude that the differential network $\Delta$ is one entry of the canonical parameter for this distribution. When using vanilla MLE to learn this exponential distribution (i.e., estimating canonical parameter), the backward mapping of $\Delta$ can be easily inferred from the two sample covariance matrices using $(\widehat{\Sigma}_d^{-1} - \widehat{\Sigma}_c^{-1})(Section\ A.1)$.

**(2) Proxy Backward Mapping:** Now the key is to find a closed-form and statistical guaranteed estimator as proxy backward mapping of $\Delta$ under high-dimensional cases. Inspired by the elementary estimator for sGGM, we choose $[T_v(\widehat{\Sigma}_d)]^{-1} - [T_v(\widehat{\Sigma}_c)]^{-1}$ as the proxy backward mapping for $\Delta$. Here

$$[T_v(A)]_{ij} := \rho_v(A_{ij}) \quad (2.10)$$

where $\rho_v(\cdot)$ is chosen to be a soft-thresholding function. We therefore obtain the following DIFFEE objective function for estimating sparse changes in GGM structure:

$$\operatorname*{argmin}_{\Delta} ||\Delta||_1$$

Subject to: $||\Delta - \left([T_v(\widehat{\Sigma}_d)]^{-1} - [T_v(\widehat{\Sigma}_c)]^{-1}\right)||_\infty \leq \lambda_n$ (2.11)

Here $\lambda_n > 0$ is the tuning parameter.

The optimization in Eq. (2.11) seeks an estimator with minimum complexity with regard to the $\ell_1$ regularization, at the same time being close enough to the 'initial estimator' $[T_v(\widehat{\Sigma}_d)]^{-1} - [T_v(\widehat{\Sigma}_c)]^{-1}$ according to the element-wise $\ell_\infty$ norm. This formulation ensures that the final estimator (solution of Eq. (2.11)) has the desired sparse structure.

Theoretically, the choice of $\ell_1$ and $\ell_\infty$ in Eq. (2.9) connects to the asymptotic error bounds of the final estimators. In Section 2.5, we theoretically prove that the statistical convergence rate of DIFFEE achieves the same sharp convergence rate as the state-of-the-art estimators for differential network. Our proofs are inspired by the unified framework of the high-dimensional statistics[20] and EE for sGGM[31].

[31] proved that when (p>n), the proxy backward mapping $[T_v(\widehat{\Sigma})]^{-1}$ in their EE-sGGM achieves the sharp convergence rate to its truth (i.e., by proving $||T_v(\widehat{\Sigma}))^{-1} - \Sigma^{*-1}||_\infty = O(\sqrt{\frac{\log p}{n}})$). The proof was extended from the previous study [26] who devised $T_v(\widehat{\Sigma})$ for estimating covariance matrix consistently under high-dimensional cases. We use the convergence results from [26] and [31] in Section 2.5 for deriving the statistical convergence rates of DIFFEE (details in Section A.2).

**(3) Closed Form Solution:** To solve Eq. (2.11), we get the following closed form solution:
$$\widehat{\Delta} = S_{\lambda_n}([T_v(\widehat{\Sigma}_d)]^{-1} - [T_v(\widehat{\Sigma}_c)]^{-1}) \quad (2.12)$$
Where $[T_v(\widehat{\Sigma}_d)]^{-1} - [T_v(\widehat{\Sigma}_c)]^{-1}$ is the pre-computed proxy backward mapping. Here $[S_\lambda(A)]_{ij} = \text{sign}(A_{ij})\max(|A_{ij}|-\lambda, 0)$ is the same soft-thresholding function in Eq. (2.5). Algorithm 1 shows the detailed steps of the DIFFEE estimator. Being non-iterative, the closed form solution helps DIFFEE achieve significant computational advantages over other estimators.

---
**Algorithm 1** DIFFEE
---
**input** Two data matrices $\mathbf{X}_c$ and $\mathbf{X}_d$.
**input** Hyper-parameter: $\lambda_n$ and $v$
**output** $\Delta$
1: Compute $[T_v(\widehat{\Sigma}_c)]^{-1}$ and $[T_v(\widehat{\Sigma}_d)]^{-1}$ from $\widehat{\Sigma}_c$ and $\widehat{\Sigma}_d$.
2: Compute $\Delta = S_{\lambda_n}([T_v(\widehat{\Sigma}_d)]^{-1} - [T_v(\widehat{\Sigma}_c)]^{-1})$
**output** $\Delta$
---

### 2.4 Analysis of Computational Complexity

The closed form solution (Eq. (2.12)) brings significant advantages in hyper-parameter tuning. This is because we only need to compute the proxy backward mapping $[T_v(\widehat{\Sigma}_d)]^{-1} - [T_v(\widehat{\Sigma}_c)]^{-1}$ once. Then the model selection just executes a fast and simple element-wise soft-thresholding operator using different values of hyper-parameter $\lambda_n$ ( Eq. (2.12)).

In details, DIFFEE includes four non-iterative operations in its computation:

1. Estimating two covariance matrices. The computational complexity is $O(\max(n_c, n_d)p^2)$.
2. The element-wise soft-thresholding operations $[T_v(\cdot)]$, that cost $O(p^2)$.

Table 1: Compare the asymptotic time complexity. DIFFEE is the best among all the estimators. Here $T$ is the number of iterations.

| DIFFEE | FusedGLasso | Density Ratio | Diff-CLIME |
|---|---|---|---|
| $O(p^3)$ | $O(T*p^3)$ | $O((n_c + p^2)^3)$ | $O(p^8)$ |

3. The matrix inversions [2] $[T_v(\cdot)]^{-1}$ to get the proxy backward mapping, that cost $O(p^3)$.
4. The element-wise soft-thresholding operation $S_{\lambda_n}$ that costs $O(p^2)$.

Therefore, the total asymptotic computational complexity of DIFFEE estimator is $O(p^3)$.

In Table 1, we compare the asymptotic computational complexity of our method to the baselines. DIFFEE achieves the best computational complexity compared to the state-of-the-art baselines. This is because:

- All existing estimators for differential network estimation have used an iterative optimization procedure to find the solution. In each iteration, their estimations require at least $O(p^3)$ computational cost.
- For tuning the sparsity hyperparameter $\lambda_n$, DIFFEE only needs to re-run its element-wise soft-thresholding operation $S_{\lambda_n}$ that cost $O(p^2)$. In contrast, all the baselines have to re-run the whole algorithm for each value of the hyper-parameter $\lambda_n$.
- Most estimators have two hyperparameters for tuning. FusedGlasso (Eq. (2.6)) and DensityRatio (Eq. (2.8)) both need to tune the hyperparameter $\lambda_2$ [3]. Both tuning are much more expensive than DIFFEE in computation. DIFFEE needs to tune the hyperparamter $v$, but it costs only $O(p^2)$.
- Diff-CLIME has one hyperparameter $\lambda_n$ for tuning, however, its asymptotic time cost ($O(p^8)$) is significantly more demanding than DIFFEE [4]. In summary, Diff-CLIME can not handle large-scale cases, like $p > 100$. For example, in our experiments Diff-CLIME can not even finish on a case of $p = 200$ after two days of running.

### 2.5 Strong Statistical Guarantees of DIFFEE

In this section, we provide a statistical convergence analysis of DIFFEE Eq. (2.9) under the following struc-

---
[2] Many faster algorithms exist for speeding up matrix inversion and matrix multiplication. The best known asymptotic cost of matrix inversion is $O(p^{2.373})$ (Wikipedia). Besides both operations can be further improved up by paralelization

[3] The optimization problem of DensityRatio is a quadratic programming problem with $n_c + p^2$ variables. Based on the result from [4], the computational complexity of quadratic problem with $b$ variables is $O(b^3)$. Therefore, the time complexity of DensityRatio is $O((n_c + p^2)^3)$.

[4] The optimization problem of Diff-CLIME is a linear programming problem with $p^2$ variables. Based on the result from [6], the computational complexity of linear problem with $b$ variables is $O(b^4)$. Therefore, the time complexity of Diff-CLIME is $O((p^2)^4)$.

Fast and Scalable Learning of Sparse Changes in High-Dimensional Gaussian Graphical Model Structure 6tural assumption:

**(C-Sparsity):** The 'true' canonical exponential family parameter for $\Delta^*$ (sparse change between two GGM structures) is exactly sparse with $k$ non-zero entries indexed by a supported set $S$. All other elements equal to 0 (in $S^c$).

**Theorem 2.1.** *Consider any differential network in Eq. (1.1) whose sparse canonical parameter $\Delta^*$ satisfies the **(C-Sparsity)** assumption. Suppose we compute the solution of Eq. (2.9) with a bounded $\lambda_n$ such that $\lambda_n \geq ||\Delta^* - \mathcal{B}^*(\widehat{\Sigma}_d, \widehat{\Sigma}_c)||_\infty$, then the optimal solution $\widehat{\Delta}$ satisfies the following error bounds:*

$$||\widehat{\Delta} - \Delta^*||_\infty \leq 2\lambda_n$$
$$||\widehat{\Delta} - \Delta^*||_F \leq 4\sqrt{k}\lambda_n \qquad (2.13)$$
$$||\widehat{\Delta} - \Delta^*||_1 \leq 8k\lambda_n$$

*Proof.* See detailed proof in Section A.2.2 □

Theorem (2.1) provides a general bound for any selection of $\lambda_n$ and $\mathcal{B}^*(\widehat{\Sigma}_d, \widehat{\Sigma}_c)$. We then use Theorem (2.1) to derive the statistical convergence rate of DIFFEE whose choice of the proxy backward mapping is $\mathcal{B}^*(\widehat{\Sigma}_d, \widehat{\Sigma}_c) = [T_v(\widehat{\Sigma}_d)]^{-1} - [T_v(\widehat{\Sigma}_c)]^{-1}$. This gives us the following corollary:

**Corollary 2.2.** *Suppose the high-dimensional setting, i.e., $p > \max(n_c, n_d)$. Let $v := a\sqrt{\frac{\log p}{\min(n_c, n_d)}}$. Then for $\lambda_n := \frac{8\kappa_1 a}{\kappa_2}\sqrt{\frac{\log p}{\min(n_c, n_d)}}$ and $\min(n_c, n_d) > c\log p$, with a probability of at least $1 - 2C_1\exp(-C_2Kp\log(Kp))$, the estimated optimal solution $\widehat{\Delta}$ has the following error bound:*

$$||\widehat{\Delta} - \Delta^*||_\infty \leq \frac{16\kappa_1 a}{\kappa_2}\sqrt{\frac{\log p}{\min(n_c, n_d)}}$$
$$||\widehat{\Delta} - \Delta^*||_F \leq \frac{32\kappa_1 a}{\kappa_2}\sqrt{\frac{k\log p}{\min(n_c, n_d)}} \qquad (2.14)$$
$$||\widehat{\Delta} - \Delta^*||_1 \leq \frac{64\kappa_1 a}{\kappa_2}k\sqrt{\frac{\log p}{\min(n_c, n_d)}}$$

*where $a$, $c$, $\kappa_1$ and $\kappa_2$ are constants.*

*Proof.* See detailed proof in Section A.2.4 (especially from Eq. (A.31) to Eq. (A.36)). □

DIFFEE has achieved the same convergence rates as the Diff-CLIME[35] and the DensityRatio estimator [16]. The FusedGLasso estimator has not provided such convergence rate analysis.

To derive the statistical error bound of DIFFEE, we need to assume that $[T_v(\widehat{\Sigma}_c)]^{-1}$ and $[T_v(\widehat{\Sigma}_d)]^{-1}$ are well-defined. This is ensured by assuming that the true $\Omega_c^*$ and $\Omega_d^*$ satisfy the following conditions [31]:

**(C-MinInf-$\Sigma$):** The true $\Omega_c^*$ and $\Omega_d^*$ of Eq. (1.1) have bounded induced operator norm, i.e., $|||\Omega_c^*|||_\infty := \sup_{w \neq 0 \in \mathbb{R}^p} \frac{||\Sigma_c^* w||_\infty}{||w||_\infty} \leq \kappa_1$ and $|||\Omega_d^*|||_\infty := \sup_{w \neq 0 \in \mathbb{R}^p} \frac{||\Sigma_d^* w||_\infty}{||w||_\infty} \leq \kappa_1$.

**(C-Sparse-$\Sigma$):** The two true covariance matrices $\Sigma_c^*$ and $\Sigma_d^*$ are "approximately sparse" (following [3]). For some constant $0 \leq q < 1$ and $c_0(p)$, $\max_i \sum_{j=1}^{p} |[\Sigma_c^*]_{ij}|^q \leq c_0(p)$ and $\max_i \sum_{j=1}^{p} |[\Sigma_d^*]_{ij}|^q \leq c_0(p)$. [5]

We additionally require $\inf_{w \neq 0 \in \mathbb{R}^p} \frac{||\Omega_c^* w||_\infty}{||w||_\infty} \geq \kappa_2$ and $\inf_{w \neq 0 \in \mathbb{R}^p} \frac{||\Omega_d^* w||_\infty}{||w||_\infty} \geq \kappa_2$.

## 3 Experiments

We use two models of simulated datasets as well as a real world dataset for empirical comparisons.

- The first model mimics real world networks with a sparse differential network containing only hub nodes. This model can evaluate whether the method can efficiently infer the hub nodes in the differential network or not. In [35], the authors claim that if the change estimator also assumes the sparsity structure in $\Omega_c$ and $\Omega_d$, then the estimator cannot achieve a good result on datasets generated by this data model.
- The second data simulation model, in contrast, generates random graphs that differ by a sparse random differential network. It evaluates the estimation performance of a certain estimator for inferring the randomly-generated differential networks.
- The real world dataset is a human brain fMRI dataset with two groups of subjects: autism and control. Our choice of this dataset is motivated by the recent literature in neuroscience that has suggested functional networks are not sparse. On the other hand, differences in functional connections across subjects should be sparse [2].

The two simulation models allow for a thorough evaluation of DIFFEE vs the baseline methods. The real-world data allows us to compare DIFFEE versus the baselines through classification using the estimated differential graph.

### 3.1 Experimental Setup

**Baselines:** We compare DIFFEE with (1) Fused-GLasso [9], (2) DensityRatio [17], and (3) Diff-CLIME [35].

**Evaluation metrics:** We evaluate DIFFEE and the baseline methods on F1-score and running time cost.

---

[5] This indicates for some positive constant $d$, $[\Sigma_c^*]_{jj} \leq d$ and $[\Sigma_d^*]_{jj} \leq d$ for all diagonal entries. Moreover, if $q = 0$, then this condition reduces to $\Sigma_d^*$ and $\Sigma_c^*$ being sparse.

More details in Section B.

**Hyper-parameters:** We need to tune the value of three hyper-parameters in these experiments: $v$, $\lambda_n$ and $\lambda_2$. In detail:

- $v$ is used for soft-thresholding in DIFFEE. We choose $v$ from the set $\{0.001i | i = 1, 2, \ldots, 1000\}$ and pick a value that makes $T_v(\Sigma_c)$ and $T_v(\Sigma_d)$ invertible.
- $\lambda_n$ is the main hyper-parameter that control the sparsity of the estimated differential network. Based on our convergence rate analysis in Section 2.5, $\lambda_n \geq C\sqrt{\frac{\log p}{\min(n_c, n_d)}}$. Accordingly, we choose $\lambda_n$ from a range of $\{0.01 \times \sqrt{\frac{\log p}{\min(n_c, n_d)}} \times i | i \in \{1, 2, 3, \ldots, 30\}\}$. The $\lambda_n$ in the DensityRatio is tuned by their package.
- $\lambda_2$ controls individual graph's sparsity in Fused-GLasso. We choose $\lambda_1 = 0.0001$ (a very small value) for all experiments to ensure only the differential network is sparse. $\lambda_2$ in the DensityRatio is set to 0.2 according to their package.

**Two models to generate simulated datasets:** Using the following two graph models, we generate multiple sets of synthetic multivariate-Gaussian datasets.

- **Model 1 – mimic real-world networks with hub nodes:** Inspired by [35], this model assumes that the graphs mimic real-world networks [21]. We first generate $\Omega_d$ as a network with $s \cdot \frac{p(p-1)}{2}$ edges following a power-law degree distribution with an expected power parameter of 2. Here $s$ is a parameter that controls the sparsity of the two graphs. A larger value of $s$ corresponds to denser graphs. Next, the value of each nonzero entry of $\Omega_d$ is generated from a uniform distribution with $[-10/p, -4/p] \cup [4/p, 10/p]$, where division by $p$ ensures the positive definiteness of $\Omega_c$ and $\Omega_d$. The diagonals are then set to 1 and $\Omega_d$ is symmetrized by averaging it with its transpose ($\frac{1}{2}(\Omega_d + \Omega_d^T)$). The differential network $\Delta$ is generated by the top 20% edges of the top 2 hub nodes in $\Omega_d$. $\Omega_c = \Omega_d - \Delta$.
- **Model 2 – random graph model:** Following [25], this model assumes $\Omega_c = \mathbf{B}_c + \mathbf{B}_S + \delta_c I$ and $\Omega_d = \mathbf{B}_d + \mathbf{B}_S + \delta_d I$, where each off-diagonal entry in $\mathbf{B}_c$ and $\mathbf{B}_d$ are generated independently and equals 0.5 with probability 0.1 and 0 with probability 0.9. The shared part $\mathbf{B}_S$ is generated independently and equal to 0.5 with probability $0.1s$ and 0 with probability $1 - 0.1s$. Similar to Model 1, $s$ controls the sparsity of the two graphs. $\delta_c$ and $\delta_d$ are selected large enough to guarantee the positive definiteness. A clear differential network structure $\Delta = \mathbf{B}_d - \mathbf{B}_c$ exists between these two graphs.

Following Model 1 or Model 2, for each case of simulated data generation, we generate two blocks of data samples following the distribution $N(0, (\Omega_c)^{-1})$ and $N(0, (\Omega_d)^{-1})$. Details see Section B.

### 3.2 Experiments on Simulated Datasets

**Experimental Design:** By varying the number of features $p$, amount of sparsity $s$, and the number of samples ($n_c, n_d$), we can generate many cases of simulated datasets. This allows us to comprehensively evaluate DIFFEE across a wide range of data situations. To this end, we design the following three sets of synthetic experiments by varying $p$, $s$, $n_c$, and $n_d$:

- $p$ (the number of features): The first set of experiments varies $p$ in the set of $\{50, 100, 200, 300, 400, 500\}$ while setting $n_c$ and $n_d$ as $p/2$ and the sparsity parameter $s = 0.2$.
- $s$ (the sparsity): In the second set of experiments, we vary the value of the sparsity parameter $s$ in the set of $\{0.1, 0.2, \ldots, 0.7\}$, while using $p = 200$ and $n_c = n_d = p/2$.
- $n_c$ and $n_d$ (the number of samples): In the third set, we vary the number of samples in both groups and set $p = 200$ and $s = 0.2$. We group this set of experiments into two categories: low-dimensional cases, and high-dimensional cases. For the high dimensional case, we vary $n_c$ and $n_d$ from the value set of $\{p/4, p/2\}$. Similarly, for the low dimensional case, we vary $n_c$ and $n_d$ from the value set of $\{p, 2p, 3p\}$.

**Experiment Results:** We compare DIFFEE with the baselines regarding two aspects– (a) Effectiveness, and (b) Scalability.

Figure 2: F1-Score of DIFFEE vs the F1-Score of the best performing baseline. The more points below the diagonal line, the better. (a) On simulated datasets from Model 1 (b) On simulated datasets from Model 2. (Black up-triangles describe 'varying $(n_c, n_d)$ in low dimensions'; Black down-triangles describe 'varying $(n_c, n_d)$ in high-dimensions; Red diamonds represent 'varying $s$'; and Blue stars represent 'varying $p$'.)

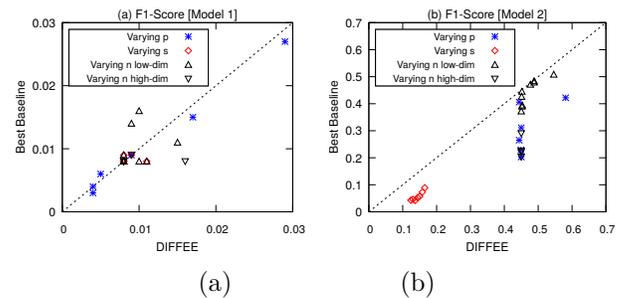

(a)      (b)

**(a) Effectiveness:** We evaluate the prediction effectiveness of using F1-Score. Figure 2 presents the summarized results of our DIFFEE versus baselines on all 50 cases of simulated datasets. As explained above, the simulated datasets are generated by varying the parameters $p$, $s$, $n_c$, and $n_d$ by data Model 1 and Model 2. In Figure 2 (a) and (b), we plot the F1-Score of DIFFEE vs the F1-Score of the best performing baseline on each simulated case from Model 1 and Model 2, respectively. Each point in the two figures is obtained by comparing DIFFEE vs. the best baseline among all baselines on one simulated case. Each point below the line $y = x$ indicates that DIFFEE achieves better performance over baselines. Overall Figure 2 shows that DIFFEE outperforms the corresponding best baseline in almost all cases. The only two points for which DIFFEE doesn't do as well as the corresponding best baseline DensityRatio are two low dimensional cases. This is as expected because the design of DIFFEE is for high-dimensional cases (i.e., the choices of proxy backward mapping). Details of F1-Scores from all simulation cases and discussions of low F1 values on Model 1 are in Appendix.

Figure 3: Time Cost(log(seconds)) of DIFFEE versus the baseline methods (a) Time vs. number of features(p) for Model 1. (b)Time vs. number of features(p) for Model 2. (c) Time vs. sparsity(s) for Model 1. (d) Time vs. sparsity(s) for Model 2. (e)Time vs. number of samples in 'c' case ($n_c$) for Model 1. (f) Time vs. number of samples in the 'c' case ($n_c$) for Model 2.

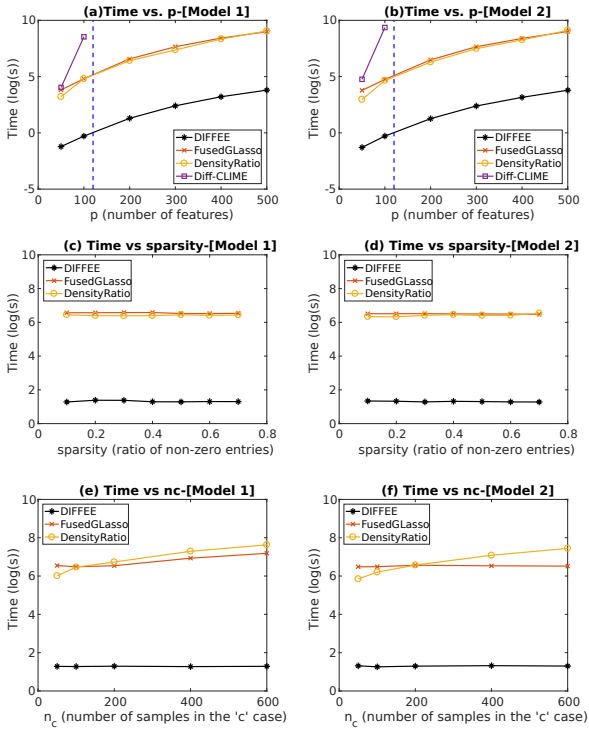

**(b) Scalability**: To evaluate DIFFEE and the baselines on scalability, Figure 3 presents the time cost vs. varying $p$, varying sparsity ($s$) and varying number of samples in the 'c' group ($n_c$). Figure 3 (a),(c) and (e) show time results from data Model 1. Figure 3 (b),(d) and (f) correspond to datasets from Model 2. We interpolate the points of computation time from each estimator into curves. For each simulation case, the computation time for each estimator is the summation of a method's execution time over all values of $\lambda_n$. Figure 3 shows that in general the time costs of FusedGLasso and DensityRatio are roughly comparable. DIFFEE is about 100 times better than both (detailed numbers are provided in Table 3 to Table 10). Diff-CLIME is extremely slow when $p$ increases. Because Figure 3 (c),(d),(e) and (f) are about data cases with $p = 200$, we can not run Diff-CLIME on these cases (it cannot finish any $p = 200$ case for a single value of $\lambda_n$ by one day). Interestingly, the empirical time results match the computational analysis in Table 1. Especially DensityRatio's time cost grows quickly when $n_c$ increases. In contrast the running time of DIFFEE and FusedGLasso are not connected strongly to the size of samples. Overall DIFFEE costs much less computation time than the baselines and can significantly scale up to larger $p$.

### 3.3 A Real-World Dataset about Functional Connectivity among Brain Regions

We then use DIFFEE for a classification task on a well-known human brain fMRI dataset: ABIDE[10].

**ABIDE Dataset:** This data is from the Autism Brain Imaging Data Exchange (ABIDE) [10], a publicly available resting-state fMRI dataset. The ABIDE data aims to understand human brain connectivity and how it reflects neural disorders [27]. The data is retrieved from the Preprocessed Connectomes Project [7], where preprocessing is performed using the Configurable Pipeline for the Analysis of Connectomes (CPAC) [8] without global signal correction or band-pass filtering. After preprocessing with this pipeline, 871 individuals remain (468 diagnosed with autism). Signals for the 160 (number of features $p = 160$) regions of interest (ROIs) in the often-used Dosenbach Atlas [11] are examined.

**Cross-validation:** Classification is performed using the 3-fold cross-validation suggested by the literature [22][28]. The subjects are randomly partitioned into three equal sets: a training set, a validation set, and a test set. Each estimator produces $\hat{\Delta}$ using the training set. Then, these differential networks are used as inputs to linear discriminant analysis (LDA), which is tuned via cross-validation on the validation set. Finally, accuracy is calculated by running LDA on the test set. This classification process aims to assess the ability of an estimator to learn the differential patterns of the connectome structures. Notably, the DensityRatio method cannot be compared on this data, because the method does not provide the precision matrices necessary for LDA.

**Classification Results:** Table 2 displays the maximum accuracy achieved by DIFFEE, FusedGLasso, and Diff-CLIME, after tuning over hyperparameters. DIFFEE yields a classification accuracy of 57.58% distinguishing the autism and control groups, outperforming the FusedGLasso and Diff-CLIME estimators.

Table 2: Classification accuracy obtained on the ABIDE dataset using DIFFEE, FusedGLasso, and Diff-CLIME. DIFFEE achieves the highest classification accuracy.

| Method | DIFFEE | FusedGLasso | Diff-CLIME |
|---|---|---|---|
| Accuracy (%) | **57.58%** | 56.90% | 53.79% |

## 4 Conclusion

This paper proposes a simple closed-form estimator, DIFFEE for learning sparse change between two GGM structures. DIFFEE can scale up to large-scale settings ($p > 100$) and achieves the same asymptotic convergence rate as previous estimators. Empirically DIFFEE improves the state-of-the-art with better F1-scores and cheaper time cost (about 100 times faster).

## A  Appendix of Method

### A.1  Backward mapping for Exponential Families

The solution of vanilla graphical model MLE can be expressed as a backward mapping[29] for an exponential family distribution. It estimates the model parameters (canonical parameter $\theta$) from certain (sample) moments. We provide detailed explanations about backward mapping of exponential families, backward mapping for Gaussian special case and backward mapping for differential network of GGM in this section.

**Backward mapping:** Essentially the vanilla graphical model MLE can be expressed as a backward mapping that computes the model parameters corresponding to some given moments in an exponential family distribution. For instance, in the case of learning GGM with vanilla MLE, the backward mapping is $\widehat{\Sigma}^{-1}$ that estimates $\Omega$ from the sample covariance (moment) $\widehat{\Sigma}$.

Suppose a random variable $X \in \mathbb{R}^p$ follows the exponential family distribution:

$$\mathbb{P}(X;\theta) = h(X)\exp\{<\theta,\phi(\theta)> -A(\theta)\} \quad (A.1)$$

Where $\theta \in \Theta \subset \mathbb{R}^d$ is the canonical parameter to be estimated and $\Theta$ denotes the parameter space. $\phi(X)$ denotes the sufficient statistics as a feature mapping function $\phi : \mathbb{R}^p \to \mathbb{R}^d$, and $A(\theta)$ is the log-partition function. We then define mean parameters $v$ as the expectation of $\phi(X)$: $v(\theta) := \mathbb{E}[\phi(X)]$, which can be the first and second moments of the sufficient statistics $\phi(X)$ under the exponential family distribution. The set of all possible moments by the moment polytope:

$$\mathcal{M} = \{v | \exists p \text{ is a distribution s.t. } \mathbb{E}_p[\phi(X)] = v\} \quad (A.2)$$

Mostly, the graphical model inference involves the task of computing moments $v(\theta) \in \mathcal{M}$ given the canonical parameters $\theta \in H$. We denote this computing as **forward mapping**:

$$\mathcal{A} : H \to \mathcal{M} \quad (A.3)$$

The learning/estimation of graphical models involves the task of the reverse computing of the forward mapping, the so-called **backward mapping** [29]. We denote the interior of $\mathcal{M}$ as $\mathcal{M}^0$. **backward mapping** is defined as:

$$\mathcal{A}^* : \mathcal{M}^0 \to H \quad (A.4)$$

which does not need to be unique. For the exponential family distribution,

$$\mathcal{A}^* : v(\theta) \to \theta = \nabla A^*(v(\theta)). \quad (A.5)$$

Where $A^*(v(\theta)) = \sup_{\theta \in H} <\theta, v(\theta)> -A(\theta)$.

**Backward Mapping: Gaussian Case** If a random variable $X \in \mathbb{R}^p$ follows the Gaussian Distribution $N(\mu, \Sigma)$. then $\theta = (\Sigma^{-1}\mu, -\frac{1}{2}\Sigma^{-1})$. The sufficient statistics $\phi(X) = (X, XX^T)$, $h(x) = (2\pi)^{-\frac{k}{2}}$, and the log-partition function

$$A(\theta) = \frac{1}{2}\mu^T \Sigma^{-1}\mu + \frac{1}{2}\log(|\Sigma|) \quad (A.6)$$

When performing the inference of Gaussian Graphical Models, it is easy to estimate the mean vector $v(\theta)$, since it equals to $\mathbb{E}[X, XX^T]$.

When learning the GGM, we estimate its canonical parameter $\theta$ through vanilla MLE. Because $\Sigma^{-1}$ is one entry of $\theta$ we can use the backward mapping to estimate $\Sigma^{-1}$.

$$\theta = (\Sigma^{-1}\mu, -\frac{1}{2}\Sigma^{-1}) = \mathcal{A}^*(v) = \nabla A^*(v)$$
$$= ((\mathbb{E}_\theta[XX^T] - \mathbb{E}_\theta[X]\mathbb{E}_\theta[X]^T)^{-1}\mathbb{E}_\theta[X], \quad (A.7)$$
$$-\frac{1}{2}(\mathbb{E}_\theta[XX^T] - \mathbb{E}_\theta[X]\mathbb{E}_\theta[X]^T)^{-1}).$$

By plugging in Eq. (A.6) into Eq. (A.5), we get the backward mapping of $\Omega$ as $(\mathbb{E}_\theta[XX^T] - \mathbb{E}_\theta[X]\mathbb{E}_\theta[X]^T)^{-1}) = \widehat{\Sigma}^{-1}$, easily computable from the sample covariance matrix.

### A.1.1 Backward Mapping for Differential Network of Two GGMs

When the random variables $X_c, X_d \in \mathbb{R}^p$ follows the Gaussian Distribution $N(\mu_c, \Sigma_c)$ and $N(\mu_d, \Sigma_d)$, their density ratio (defined by [17]) essentially is a distribution in exponential families:

$$r(x, \Delta) = \frac{p_d(x)}{p_c(x)}$$
$$= \frac{\sqrt{\det(\Sigma_c)} \exp\left(-\frac{1}{2}(x-\mu_d)^T\Sigma_d^{-1}(x-\mu_d)\right)}{\sqrt{\det(\Sigma_d)} \exp\left(-\frac{1}{2}(x-\mu_c)^T\Sigma_c^{-1}(x-\mu_c)\right)}$$
$$= \exp(-\frac{1}{2}(x-\mu_d)^T\Sigma_d^{-1}(x-\mu_d)$$
$$+ \frac{1}{2}(x-\mu_c)^T\Sigma_c^{-1}(x-\mu_c)$$
$$- \frac{1}{2}(\log(\det(\Sigma_d)) - \log(\det(\Sigma_c))))$$
$$= \exp\left(-\frac{1}{2}\Delta x^2 + \mu_\Delta x - A(\mu_\Delta, \Delta)\right)$$
$$(A.8)$$

Here $\Delta = \Sigma_d^{-1} - \Sigma_c^{-1}$ and $\mu_\Delta = \Sigma_d^{-1}\mu_d - \Sigma_c^{-1}\mu_c$.

The log-partition function

$$A(\mu_\Delta, \Delta) = \frac{1}{2}\mu_d^T\Sigma_d^{-1}\mu_d - \frac{1}{2}\mu_c^T\Sigma_c^{-1}\mu_c +$$
$$\frac{1}{2}\log(\det(\Sigma_d)) - \frac{1}{2}\log(\det(\Sigma_c)) \quad (A.9)$$

The canonical parameter

$$\theta = \left(\Sigma_d^{-1}\mu_d - \Sigma_c^{-1}\mu_c, -\frac{1}{2}(\Sigma_d^{-1} - \Sigma_c^{-1})\right)$$
$$= \left(\Sigma_d^{-1}\mu_d - \Sigma_c^{-1}\mu_c, -\frac{1}{2}(\Delta)\right) \quad (A.10)$$

The sufficient statistics $\phi([X_c, X_d])$ and the log-partition function $A(\theta)$:

$$\phi([X_c, X_d]) = ([X_c, X_d], [X_c X_c^T, X_d X_d^T])$$
$$A(\theta) = \frac{1}{2}\mu_d^T\Sigma_d^{-1}\mu_d - \frac{1}{2}\mu_c^T\Sigma_c^{-1}\mu_c + \quad (A.11)$$
$$\frac{1}{2}\log(\det(\Sigma_d)) - \frac{1}{2}\log(\det(\Sigma_c))$$

And $h(x) = 1$.

Now we can estimate this exponential distribution ($\theta$) through vanilla MLE. By plugging Eq. (A.11) into Eq. (A.5), we get the following backward mapping via the conjugate of the log-partition function:

$$\theta = \left(\Sigma_d^{-1}\mu_d - \Sigma_c^{-1}\mu_c, -\frac{1}{2}(\Sigma_d^{-1} - \Sigma_c^{-1})\right)$$
$$= \mathcal{A}^*(v) = \nabla A^*(v) \quad (A.12)$$

The mean parameter vector $v(\theta)$ includes the moments of the sufficient statistics $\phi()$ under the exponential distribution. It can be easily estimated through $\mathbb{E}[([X_c, X_d], [X_c X_c^T, X_d X_d^T])]$.

Therefore the backward mapping of $\theta$ becomes,

$$\widehat{\theta} = (((\mathbb{E}_\theta[X_d X_d^T] - \mathbb{E}_\theta[X_d]\mathbb{E}_\theta[X_d]^T)^{-1}\mathbb{E}_\theta[X_d]$$
$$- (\mathbb{E}_\theta[X_c X_c^T] - \mathbb{E}_\theta[X_c]\mathbb{E}_\theta[X_c]^T)^{-1}\mathbb{E}_\theta[X_c]),$$
$$- \frac{1}{2}((\mathbb{E}_\theta[X_d X_d^T] - \mathbb{E}_\theta[X_d]\mathbb{E}_\theta[X_d]^T)^{-1} -$$
$$(\mathbb{E}_\theta[X_c X_c^T] - \mathbb{E}_\theta[X_c]\mathbb{E}_\theta[X_c]^T)^{-1})).$$
$$(A.13)$$

Because the second entry of the canonical parameter $\theta$ is $(\Sigma_d^{-1} - \Sigma_c^{-1})$, we get the backward mapping of $\Delta$ as

$$((\mathbb{E}_\theta[X_d X_d^T] - \mathbb{E}_\theta[X_d]\mathbb{E}_\theta[X_d]^T)^{-1}$$
$$- (\mathbb{E}_\theta[X_c X_c^T] - \mathbb{E}_\theta[X_c]\mathbb{E}_\theta[X_c]^T)^{-1}) \quad (A.14)$$
$$= \widehat{\Sigma}_d^{-1} - \widehat{\Sigma}_c^{-1}$$

This can be easily inferred from two sample covariance matrices $\widehat{\Sigma}_d$ and $\widehat{\Sigma}_c$ (Att: when under low-dimensional settings).

### A.2 Appendix:Proof

#### A.2.1 Derivation of Theorem (2.1)

DIFFEE formulation Eq. (2.11) and EE-sGGM Eq. (2.3) are special cases of the following generic formulation:

$$\operatorname*{argmin}_{\theta} \mathcal{R}(\theta)$$
$$\text{subject to:} \mathcal{R}^*(\theta - \widehat{\theta}_n) \leq \lambda_n \quad (A.15)$$

Where $\mathcal{R}^*(\cdot)$ is the dual norm of $\mathcal{R}(\cdot)$,

$$\mathcal{R}^*(v) := \sup_{u \neq 0} \frac{<u, v>}{\mathcal{R}(u)} = \sup_{\mathcal{R}(u) \leq 1} <u, v> . \quad (A.16)$$

Connecting Eq. (2.11) and Eq. (A.15), $\mathcal{R}()$ is the $\ell_1$ norm, $\mathcal{R}^*()$ is the $\ell_\infty$-norm, and $\ell_\infty$-norm is the dual norm of $\ell_1$-norm. $\widehat{\theta}_n$ represents a backward mapping (or proxy backward mapping well-defined in high-dimensional settings) of $\theta$, which is a close approximation of $\theta^*$.

Following the unified framework [20], we first decompose the parameter space into a subspace pair $(\mathcal{M}, \bar{\mathcal{M}}^\perp)$, where $\bar{\mathcal{M}}$ is the closure of $\mathcal{M}$. Here $\bar{\mathcal{M}}^\perp := \{v \in \mathbb{R}^p | < u, v > = 0, \forall u \in \bar{\mathcal{M}}\}$. $\mathcal{M}$ is the **model subspace** that typically has a much lower dimension than the original high-dimensional space. $\bar{\mathcal{M}}^\perp$ is the **perturbation subspace** of parameters. For further proofs, we assume the regularization function in Eq. (A.15) is **decomposable** w.r.t the subspace pair $(\mathcal{M}, \bar{\mathcal{M}}^\perp)$.

**(C1)** $\mathcal{R}(u+v) = \mathcal{R}(u) + \mathcal{R}(v)$, $\forall u \in \mathcal{M}, \forall v \in \bar{\mathcal{M}}^\perp$.

[20] showed that most regularization norms are decomposable corresponding to a certain subspace pair.

**Definition A.1.** *Subspace Compatibility Constant*

*Subspace compatibility constant is defined as $\Psi(\mathcal{M}, |\cdot|) := \sup_{u \in \mathcal{M}\setminus\{0\}} \frac{\mathcal{R}(u)}{|u|}$ which captures the relative value between the error norm $|\cdot|$ and the regularization function $\mathcal{R}(\cdot)$.*

For simplicity, we assume there exists a true parameter $\theta^*$ which has the exact structure w.r.t a certain subspace pair. Concretely:

**(C2)** $\exists$ a subspace pair $(\mathcal{M}, \bar{\mathcal{M}}^\perp)$ such that the true parameter satisfies $\text{proj}_{\mathcal{M}^\perp}(\theta^*) = 0$

Then we have the following theorem.

**Theorem A.2.** *Suppose the regularization function in Eq. (A.15) satisfies condition **(C1)**, the true parameter of Eq. (A.15) satisfies condition **(C2)**, and $\lambda_n$ satisfies that $\lambda_n \geq \mathcal{R}^*(\widehat{\theta}_n - \theta^*)$. Then, the optimal solution $\widehat{\theta}$ of Eq. (A.15) satisfies:*

$$\mathcal{R}^*(\widehat{\theta} - \theta^*) \leq 2\lambda_n \quad (A.17)$$

$$||\widehat{\theta} - \theta^*||_2 \leq 4\lambda_n \Psi(\bar{\mathcal{M}}) \quad (A.18)$$

$$\mathcal{R}(\widehat{\theta} - \theta^*) \leq 8\lambda_n \Psi(\bar{\mathcal{M}})^2 \quad (A.19)$$

For the proposed DIFFEE model, $\mathcal{R} = ||\cdot||_1$. Based on the results in [20], $\Psi(\bar{\mathcal{M}}) = \sqrt{k}$, where $k$ is the total number of nonzero entries in $\Delta$. Using $\mathcal{R} = ||\cdot||_1$ in Theorem (A.2), we have the following theorem (the same as Theorem (2.1)),

**Theorem A.3.** *Suppose that $\mathcal{R} = ||\cdot||_1$ and the true parameter $\Delta^*$ satisfy the conditions **(C1)(C2)** and $\lambda_n \geq \mathcal{R}^*(\widehat{\Delta} - \Delta^*)$, then the optimal point $\widehat{\Delta}$ of Eq. (2.11) has the following error bounds: $||\widehat{\Delta} - \Delta^*||_\infty \leq 2\lambda_n$, $||\widehat{\Delta} - \Delta^*||_2 \leq 4\sqrt{k}\lambda_n$, and $||\widehat{\Delta} - \Delta^*||_1 \leq$* $8k\lambda_n$

### A.2.2 Proof of Theorem (A.2)

*Proof.* Let $\delta := \widehat{\theta} - \theta^*$ be the error vector that we are interested in.

$$\begin{aligned}\mathcal{R}^*(\widehat{\theta} - \theta^*) &= \mathcal{R}^*(\widehat{\theta} - \widehat{\theta}_n + \widehat{\theta}_n - \theta^*) \\ &\leq \mathcal{R}^*(\widehat{\theta}_n - \widehat{\theta}) + \mathcal{R}^*(\widehat{\theta}_n - \theta^*) \leq 2\lambda_n\end{aligned} \quad (A.20)$$

By the fact that $\theta^*_{\mathcal{M}^\perp} = 0$, and the decomposability of $\mathcal{R}$ with respect to $(\mathcal{M}, \bar{\mathcal{M}}^\perp)$

$$\begin{aligned}&\mathcal{R}(\theta^*) \\ &= \mathcal{R}(\theta^*) + \mathcal{R}[\Pi_{\bar{\mathcal{M}}^\perp}(\delta)] - \mathcal{R}[\Pi_{\bar{\mathcal{M}}^\perp}(\delta)] \\ &= \mathcal{R}[\theta^* + \Pi_{\bar{\mathcal{M}}^\perp}(\delta)] - \mathcal{R}[\Pi_{\bar{\mathcal{M}}^\perp}(\delta)] \\ &\leq \mathcal{R}[\theta^* + \Pi_{\bar{\mathcal{M}}^\perp}(\delta) + \Pi_{\bar{\mathcal{M}}}(\delta)] + \mathcal{R}[\Pi_{\bar{\mathcal{M}}}(\delta)] \\ &\quad - \mathcal{R}[\Pi_{\bar{\mathcal{M}}^\perp}(\delta)] \\ &= \mathcal{R}[\theta^* + \delta] + \mathcal{R}[\Pi_{\bar{\mathcal{M}}}(\delta)] - \mathcal{R}[\Pi_{\bar{\mathcal{M}}^\perp}(\delta)]\end{aligned} \quad (A.21)$$

Here, the inequality holds by the triangle inequality of norm. Since Eq. (A.15) minimizes $\mathcal{R}(\widehat{\theta})$, we have $\mathcal{R}(\theta^* + \Delta) = \mathcal{R}(\widehat{\theta}) \leq \mathcal{R}(\theta^*)$. Combining this inequality with Eq. (A.21), we have:

$$\mathcal{R}[\Pi_{\bar{\mathcal{M}}^\perp}(\delta)] \leq \mathcal{R}[\Pi_{\bar{\mathcal{M}}}(\delta)] \quad (A.22)$$

Moreover, by Hölder's inequality and the decomposability of $\mathcal{R}(\cdot)$, we have:

$$\begin{aligned}||\Delta||_2^2 &= \langle \delta, \delta \rangle \leq \mathcal{R}^*(\delta)\mathcal{R}(\delta) \leq 2\lambda_n \mathcal{R}(\delta) \\ &= 2\lambda_n[\mathcal{R}(\Pi_{\bar{\mathcal{M}}}(\delta)) + \mathcal{R}(\Pi_{\bar{\mathcal{M}}^\perp}(\delta))] \leq 4\lambda_n \mathcal{R}(\Pi_{\bar{\mathcal{M}}}(\delta)) \\ &\leq 4\lambda_n \Psi(\bar{\mathcal{M}})||\Pi_{\bar{\mathcal{M}}}(\delta)||_2\end{aligned} \quad (A.23)$$

where $\Psi(\bar{\mathcal{M}})$ is a simple notation for $\Psi(\bar{\mathcal{M}}, ||\cdot||_2)$.

Since the projection operator is defined in terms of $||\cdot||_2$ norm, it is non-expansive: $||\Pi_{\bar{\mathcal{M}}}(\Delta)||_2 \leq ||\Delta||_2$. Therefore, by Eq. (A.23), we have:

$$||\Pi_{\bar{\mathcal{M}}}(\delta)||_2 \leq 4\lambda_n \Psi(\bar{\mathcal{M}}), \quad (A.24)$$

and plugging it back to Eq. (A.23) yields the error bound Eq. (A.18).

Finally, Eq. (A.19) is straightforward from Eq. (A.22) and Eq. (A.24).

$$\begin{aligned}\mathcal{R}(\delta) &\leq 2\mathcal{R}(\Pi_{\bar{\mathcal{M}}}(\delta)) \\ &\leq 2\Psi(\bar{\mathcal{M}})||\Pi_{\bar{\mathcal{M}}}(\delta)||_2 \leq 8\lambda_n \Psi(\bar{\mathcal{M}})^2.\end{aligned} \quad (A.25)$$

$\square$

### A.2.3 Useful lemma(s)

**Lemma A.4.** *(Theorem 1 of [26]). Let $\delta$ be $\max_{ij} |[\frac{X^T X}{n}]_{ij} - \Sigma_{ij}|$. Suppose that $v > 2\delta$. Then, under the conditions (C-Sparse$\Sigma$), and as $\rho_v(\cdot)$ is a soft-threshold function, we can deterministically guarantee that the spectral norm of error is bounded as follows:*

$$|||T_v(\widehat{\Sigma}) - \Sigma|||_\infty \leq 5v^{1-q}c_0(p) + 3v^{-q}c_0(p)\delta \quad (A.26)$$

**Lemma A.5.** *(Lemma 1 of [23]). Let $\mathcal{A}$ be the event that*

$$||\frac{X^T X}{n} - \Sigma||_\infty \leq 8(\max_i \Sigma_{ii})\sqrt{\frac{10\tau \log p'}{n}} \quad (A.27)$$

*where $p' := \max n, p$ and $\tau$ is any constant greater than 2. Suppose that the design matrix $X$ is i.i.d. sampled from $\Sigma$-Gaussian ensemble with $n \geq 40 \max_i \Sigma_{ii}$. Then, the probability of event $\mathcal{A}$ occurring is at least $1 - 4/p'^{\tau-2}$.*

To prove the bound of $||\Delta^* - ([T_v(\widehat{\Sigma}_d)]^{-1} - [T_v(\widehat{\Sigma}_c)]^{-1})||_\infty$, we first prove the bound of $||\Omega_c^* - [T_v(\widehat{\Sigma}_c)]^{-1}||_\infty$. In the following proof, we first derive the inequality $||\Omega_c^* - [T_v(\widehat{\Sigma}_c)]^{-1}||_\infty \leq |||[T_v(\widehat{\Sigma}_c)]^{-1}|||_\infty ||\Omega_c^*||_\infty ||T_v(\widehat{\Sigma}_c) - \Sigma_c^*||_\infty$, which is bounded by multiplication of three parts. Then we use the above Lemmas and two conditions to prove the bound of each part. Finally, we combine the three results to have the whole bound of $||\Omega_c^* - [T_v(\widehat{\Sigma}_c)]^{-1}||_\infty$.

### A.2.4 Proof of Corollary (2.2)

*Proof.* In the following proof, we first prove $||\Omega_c^* - [T_v(\widehat{\Sigma}_c)]^{-1}||_\infty \leq \lambda_{n_c}$. Here $\lambda_{n_c} = \frac{4\kappa_1 a}{\kappa_2}\sqrt{\frac{\log p'}{n_c}}$ and $p' = \max(p, n_c)$

The condition (C-Sparse$\Sigma$) and condition (C-MinInf$\Sigma$) also hold for $\Omega_c^*$ and $\Sigma_c^*$. In order to utilize Theorem (A.3) for this case, we only need to show that $||\Omega_c^* - [T_v(\widehat{\Sigma}_c)]^{-1}||_\infty \leq \lambda_{n_c}$ for the setting of $\lambda_{n_c} = \frac{4\kappa_1 a}{\kappa_2}\sqrt{\frac{\log p'}{n_c}}$:

$$\begin{aligned}
||\Omega_c^* - [T_v(\widehat{\Sigma}_c)]^{-1}||_\infty &= |||T_v(\widehat{\Sigma}_c)]^{-1}(T_v(\widehat{\Sigma}_c)\Omega_c^* - I)||_\infty \\
&\leq |||[T_v(\widehat{\Sigma}_c w)]||_\infty ||T_v(\widehat{\Sigma}_c)\Omega_c^* - I||_\infty \\
&= |||[T_v(\widehat{\Sigma}_c)]^{-1}|||_\infty ||\Omega_c^*(T_v(\widehat{\Sigma}_c) - \Sigma_c^*)||_\infty \\
&\leq |||[T_v(\widehat{\Sigma}_c)]^{-1}|||_\infty ||\Omega_c^*||_\infty ||T_v(\widehat{\Sigma}_c) - \Sigma_c^*||_\infty.
\end{aligned} \quad (A.28)$$

We first compute the upper bound of $|||[T_v(\widehat{\Sigma}_c)]^{-1}|||_\infty$. By the selection $v$ in the statement, Lemma (A.4) and Lemma (A.5) hold with probability at least $1 - 4/p'^{\tau-2}$. Armed with Eq. (A.26), we use the triangle inequality of norm and the condition (C-Sparse$\Sigma$): for any $w$,

$$\begin{aligned}
||T_v(\widehat{\Sigma}_c)w||_\infty &= ||T_v(\widehat{\Sigma}_c)w - \Sigma w + \Sigma w||_\infty \\
&\geq ||\Sigma w||_\infty - ||(T_v(\widehat{\Sigma}_c) - \Sigma)w||_\infty \\
&\geq \kappa_2 ||w||_\infty - ||(T_v(\widehat{\Sigma}_c) - \Sigma)w||_\infty \\
&\geq (\kappa_2 - ||(T_v(\widehat{\Sigma}_c) - \Sigma)w||_\infty)||w||_\infty
\end{aligned} \quad (A.29)$$

Where the second inequality uses the condition (C-Sparse$\Sigma$). Now, by Lemma (A.4) with the selection of $v$, we have

$$|||T_v(\widehat{\Sigma}_c) - \Sigma|||_\infty \leq c_1 (\frac{\log p'}{n_c})^{(1-q)/2} c_0(p) \quad (A.30)$$

where $c_1$ is a constant related only on $\tau$ and $\max_i \Sigma_{ii}$. Specifically, it is defined as $6.5 \times (16(\max_i \Sigma_{ii})\sqrt{10\tau})^{1-q}$. Hence, as long as $n_c > (\frac{2c_1 c_0(p)}{\kappa_2})^{\frac{2}{1-q}} \log p'$ as stated, so that $|||T_v(\widehat{\Sigma}_c) - \Sigma|||_\infty \leq \frac{\kappa_2}{2}$, we can conclude that $||T_v(\widehat{\Sigma}_c)w||_\infty \geq \frac{\kappa_2}{2}||w||_\infty$, which implies $|||[T_v(\widehat{\Sigma}_c)]^{-1}|||_\infty \leq \frac{2}{\kappa_2}$.

The remaining term in Eq. (A.28) is $||T_v(\widehat{\Sigma}_c) - \Sigma_c^*||_\infty$; $||T_v(\widehat{\Sigma}_c) - \Sigma_c^*||_\infty \leq ||T_v(\widehat{\Sigma}_c) - \widehat{\Sigma}_c||_\infty + ||\widehat{\Sigma}_c - \Sigma_c^*||_\infty$. By construction of $T_v(\cdot)$ in (C-Thresh) and by Lemma (A.5), we can confirm that $||T_v(\widehat{\Sigma}_c) - \widehat{\Sigma}_c||_\infty$ as well as $||\widehat{\Sigma}_c - \Sigma_c^*||_\infty$ can be upper-bounded by $v$.

Similarly, the $[T_v(\widehat{\Sigma}_d)]^{-1}$ has the same result. Finally,

$$||\Delta^* - \left([T_v(\widehat{\Sigma}_d)]^{-1} - [T_v(\widehat{\Sigma}_c)]^{-1}\right)||_\infty \quad (A.31)$$

$$\leq ||\Omega_d - [T_v(\widehat{\Sigma}_d)]^{-1}||_\infty + ||\Omega_c - [T_v(\widehat{\Sigma}_c)]^{-1}||_\infty \quad (A.32)$$

$$\leq \frac{4\kappa_1 a}{\kappa_2}\sqrt{\frac{\log p'}{n_c}} + \frac{4\kappa_1 a}{\kappa_2}\sqrt{\frac{\log p'}{n_c}} \quad (A.33)$$

Suppose $p > \max(n_c, n_d)$, we have that
$$||\Delta^* - \left([T_v(\widehat{\Sigma}_d)]^{-1} - [T_v(\widehat{\Sigma}_c)]^{-1}\right)||_\infty \leq$$
$$\frac{8\kappa_1 a}{\kappa_2}\sqrt{\frac{\log p}{\min(n_c, n_d)}} \quad (A.34)$$

Similarly, we also have that
$$||\Delta^* - \left([T_v(\widehat{\Sigma}_d)]^{-1} - [T_v(\widehat{\Sigma}_c)]^{-1}\right)||_F \leq$$
$$\frac{32\kappa_1 a}{\kappa_2}\sqrt{\frac{k\log p}{\min(n_c, n_d)}} \quad (A.35)$$

, and
$$||\Delta^* - \left([T_v(\widehat{\Sigma}_d)]^{-1} - [T_v(\widehat{\Sigma}_c)]^{-1}\right)||_1 \leq$$
$$\frac{64\kappa_1 a}{\kappa_2} k \sqrt{\frac{\log p}{\min(n_c, n_d)}} \quad (A.36)$$

By combining all together, we can confirm that the selection of $\lambda_n$ satisfies the requirement of Theorem (A.3), which completes the proof. □

## B  Details of Experimental Setup

**Evaluation Metrics:** We evaluate DIFFEE and the baseline methods on both contexts of effectiveness and scalability.

- F1-score: We first use the edge-level F1-score to compare the predicted versus true differential graph. Here, F1 = $\frac{2 \cdot \text{Precision} \cdot \text{Recall}}{\text{Precision} + \text{Recall}}$, where Precision = $\frac{\text{TP}}{\text{TP}+\text{FP}}$ and Recall = $\frac{\text{TP}}{\text{TP}+\text{FN}}$. TP (true positive) means the number of true edges correctly estimated by the predicted differential network. FP (false positive) and FN (false negative) are the number of incorrectly predicted nonzero entries and zero entries respectively. We repeat the experiment 10 times for each method and use the average metrics for comparison. The better method achieves a higher F1-score.

- Time Cost: We use the execution time (measured in seconds or log(seconds)) for a method as a measure of its scalability. To ensure a fair comparison, we try 30 different $\lambda$ (or $\lambda_2$) and measure the total time of execution for each method. The better method uses less time[6].

- Low F1 values on Model 1 datasets: The F1-score of all cases in Figure 2(a) appear quite low. This is due to the fact that simulated differential networks from Model 1 are extremely sparse (e.g., only 0.1% non-zero edges among all possible edges). For example, if the estimated $\widehat{\Delta}$ only predicts 5% zero entries incorrectly (i.e., FP=5%) and correctly predicts all the rest entries (TP = 0.1%, TN = 94.9%). The precision equals to $\frac{\text{TP}}{\text{TP + FP}} = \frac{0.1\%}{0.1\% + 5\%} \approx 0.02$, which is a small number. The recall equals to $\frac{\text{TP}}{\text{TP + FN}} = \frac{0.1\%}{0.1\% + 0\%} = 1$. Then F1 = $\frac{\text{precision} \cdot \text{recall}}{2(\text{precision} + \text{recall})} \approx 0.01$, which is also a relatively small number. However, the estimator only wrongly inferred 5% zero entries, which is still a good result. Therefore, low F1-score doesn't mean that the estimator is bad when the differential network is extremely sparse.

  This extreme sparsity also influences other evaluation metrics. For instance, if the estimated $\widehat{\Delta}$ only includes 1% zero entries and 0.05% non-zero entries incorrectly (i.e., FP=5% and FN=0.05%) and correctly predicts all the rest entries (TP=0.05% and TN=94.9%). The TPR = $\frac{0.05\%}{0.05\% + 0.05\%} = 0.5$ and FPR = $\frac{5\%}{5\% + 94.9\%} \approx 0.2$. If you plot this point in the FPR vs. TPR curve, it is not good. However from the angle of accuracy, this method only predicts wrongly around 5% edges, which indicates that it performs well.

---

[6]The machine that we use for experiments is an Intel(R) Core(TM) i7-6850k CPU @ 3.60GHz with a 64GB memory.

Figure 4: F1-score versus Time Cost(log(seconds)) for different methods and synthetic data models (a) F1-score vs. Time for Model 1. (b)F1-score vs. Time for Model 2.

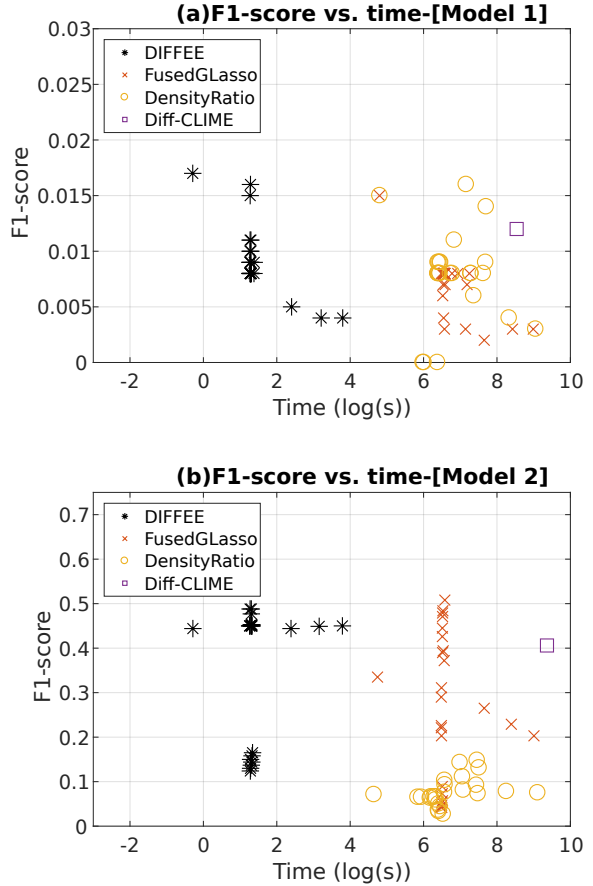

**Simulated Data Generation:** We first simulate precision matrices $\Omega_c$ and $\Omega_d$ by Model 1 or Model 2. To simulate data for the control block, we generate $n_c$ data samples following multivariate gaussian distribution with mean 0 and covariance matrix $(\Omega_c)^{-1}$. We use the multivariate distribution method from stochastic simulation [24] to sample the simulated data blocks. In our implementation, we directly use the R function "**mvrnorm**" in **MASS** package. We repeat the same process for the case group with $\Omega_d$. Then, we apply DIFFEE and baseline methods to obtain the estimated differential networks.

## C  Detailed Empirical Results

Figure 4 (a) and (b) summarize DIFFEE's better performance in both scalability and effectiveness for all experiment settings in Model 1 and Model 2, respectively. Each point in Figure4 represents both the F1-Score and Time Cost of a method. Most of the DIFFEE points lie in the top left area, indicating lesser Time Cost and higher F1-scores compared to the other baselines.

Table 3 and Table 4 present the detailed results on the simulated datasets, comparing the scalability to $p$ of our proposed method DIFFEE with the baselines FusedGLasso, Density Ratio, and Diff-CLIME. The Table 3 and Table 4 are obtained by experimental settings under Model 1 and Model 2 respectively. We vary number of features $p$ in the set of $\{100, 200, 300, 400, 500\}$. The computation time for each case is the summation of the computational time for the method over a range of $\lambda_n \in \{0.01 \times \sqrt{\frac{\log p}{\min(n_c, n_d)}} \times i | i \in \{1, 2, 3, \ldots, 30\}\}$. The F1-score for each case is the best result over a range of $\lambda_n \in \{0.01 \times \sqrt{\frac{\log p}{\min(n_c, n_d)}} \times i | i \in \{1, 2, 3, \ldots, 30\}\}$. The Diff-CLIME cannot finish any tasks in one day. So all the results in the column "Diff-CLIME" are indicated by "NA". In most of the synthetic datasets, DIFFEE achieves a higher F1-Score and less computation time than other baselines. This proves that DIFFEE outperforms the baselines in both effectiveness and scalability.

Table 5 and Table 6 present the detailed performance results of our proposed method DIFFEE and others by varying the sparsity level $s$. The Table 5 and Table 6 are obtained by Model 1 and Model 2 respectively. We vary the sparsity parameter $s$ in the set of $\{0.1, 0.2, \ldots, 0.7\}$. The computation time and F1-Score are measured similar to Table 3 and Table 4. In all of the synthetic datasets, DIFFEE performs better as indicated by its higher F1-score and lesser computation time than other baselines.

Table 7 and Table 8 present the detailed results of our proposed method–DIFFEE versus the corresponding baselines FusedGLasso, Density Ratio, and Diff-CLIME on the simulated datasets varying different $n_c$ and $n_d$ in a high-dimensional setting ($p > \max(n_c, n_d)$). The Table 7 and Table 8 are obtained by Model 1 and Model 2, respectively. We vary the number of samples $n_c$ and $n_d$ in the set of $\{p/2, p/4\}$. The computation time and F1-Score are measured similar to Table 3 and Table 4. In most of the synthetic datasets, DIFFEE achieves a higher F1-Score and less computation time than other baselines.

Table 9 and Table 10 present the performance of our proposed method–DIFFEE and other methods with varying $n_c$ and $n_d$ in a low-dimensional setting ($p > \max(n_c, n_d)$). The Table 9 and Table 10 correspond to Model 1 and Model 2, respectively. We vary the number of samples $n_c$ and $n_d$ in the set of $\{p, 2p, 3p\}$. The computation time and F1-Score are measured similar to Table 3 and Table 4. In most of the synthetic datasets, DIFFEE achieves a higher F1-Score and less computation time than other baselines.

Figure 5 and Figure 6 summarize F1-Scores for DIFFEE and the baseline methods: FusedGLasso and DensityRatio for all simulations under varying $p$, $s$ and $(n_c, n_d)$ for Model 1 and Model 2, respectively.

Table 3: Model 1 varying p

|  | Model | DIFFEE | FusedGLasso | Slower | Density Ratio | Slower | Diff-CLIME | Slower |
|---|---|---|---|---|---|---|---|---|
| F1-score | p = 50 | **0.029** | 0 | | 0.027 | | 0.016 | |
| | p = 100 | **0.017** | 0.015 | | 0.015 | | 0.012 | |
| | p = 200 | **0.009** | 0.008 | | 0.009 | | NA | |
| | p = 300 | 0.005 | 0.002 | | **0.006** | | NA | |
| | p = 400 | **0.004** | 0.003 | | 0.004 | | NA | |
| | p = 500 | **0.004** | 0.003 | | 0.003 | | NA | |
| Time (s) | p = 50 | **0.296** | 45.61 | 154× | 24.903 | 84× | 56.37 | 190× |
| | p = 100 | **0.748** | 121.537 | 162× | 122.596 | 163× | 5094.796 | 6811× |
| | p = 200 | **3.645** | 715.672 | 196× | 611.341 | 167× | NA | |
| | p = 300 | **11.064** | 2106.681 | 190× | 1584.262 | 143× | NA | |
| | p = 400 | **24.763** | 4551.419 | 183× | 4159.019 | 167× | NA | |
| | p = 500 | **44.54** | 8008.809 | 179× | 8575.529 | 192× | NA | |

Table 4: Model 2 varying p

|  | Model | DIFFEE | FusedGLasso | Slower | Density Ratio | Slower | Diff-CLIME | Slower |
|---|---|---|---|---|---|---|---|---|
| F1-score | p = 50 | **0.581** | 0.401 | | 0.082 | | 0.422 | |
| | p = 100 | **0.444** | 0.335 | | 0.071 | | 0.406 | |
| | p = 200 | **0.45** | 0.311 | | 0.066 | | NA | |
| | p = 300 | 0.444 | 0.265 | | 0.073 | | NA | |
| | p = 400 | **0.449** | 0.229 | | 0.078 | | NA | |
| | p = 500 | **0.45** | 0.203 | | 0.075 | | NA | |
| Time (s) | p = 50 | **0.274** | 43.57 | 159× | 19.35 | 70× | 116.712 | 425× |
| | p = 100 | **0.751** | 115.049 | 153× | 104.53 | 139× | 11640.82 | 15500× |
| | p = 200 | **3.528** | 657.147 | 186× | 538.842 | 152× | NA | |
| | p = 300 | **10.887** | 2106.415 | 193× | 1780.176 | 163× | NA | |
| | p = 400 | **23.462** | 4406.156 | 187× | 3859.082 | 164× | NA | |
| | p = 500 | **44.163** | 8164.19 | 184× | 9054.507 | 205× | NA | |

Table 5: Model 1 varying sparsity

|  | Model | DIFFEE | FusedGLasso | Slower | Density Ratio | Slower |
|---|---|---|---|---|---|---|
| F1-score | s = 0.1 | 0.008 | 0.003 | | **0.009** | |
| | s = 0.2 | **0.009** | 0.008 | | 0.009 | |
| | s = 0.3 | **0.008** | 0.008 | | 0.008 | |
| | s = 0.4 | **0.011** | 0.008 | | 0.008 | |
| | s = 0.5 | **0.008** | 0.006 | | 0.008 | |
| | s = 0.6 | **0.008** | 0.008 | | 0.008 | |
| | s = 0.7 | **0.008** | 0.007 | | 0.008 | |
| Time (s) | s = 0.1 | **3.606** | 712.682 | 197× | 631.582 | 175× |
| | s = 0.2 | **3.993** | 712.365 | 178× | 598.191 | 149× |
| | s = 0.3 | **3.97** | 719.859 | 181× | 595.246 | 149× |
| | s = 0.4 | **3.65** | 721.785 | 197× | 598.009 | 163× |
| | s = 0.5 | **3.632** | 679.94 | 187× | 631.062 | 173× |
| | s = 0.6 | **3.693** | 679.263 | 183× | 608.358 | 164× |
| | s = 0.7 | **3.679** | 686.979 | 186× | 624.632 | 169× |

Beilun Wang, Arshdeep Sekhon, Yanjun Qi    17

Table 6: Model 2 varying sparsity

|  | Model | DIFFEE | FusedGLasso | Slower | Density Ratio | Slower |
|---|---|---|---|---|---|---|
| F1-score | s = 0.1 | **0.165** | 0.089 | | 0.066 | |
|  | s = 0.2 | **0.158** | 0.073 | | 0.059 | |
|  | s = 0.3 | **0.15** | 0.057 | | 0.05 | |
|  | s = 0.4 | **0.144** | 0.053 | | 0.044 | |
|  | s = 0.5 | **0.137** | 0.042 | | 0.036 | |
|  | s = 0.6 | **0.13** | 0.046 | | 0.033 | |
|  | s = 0.7 | **0.124** | 0.043 | | 0.027 | |
| Time (s) | s = 0.1 | **3.817** | 671.255 | 175× | 564.679 | 147× |
|  | s = 0.2 | **3.763** | 671.499 | 178× | 559.455 | 148× |
|  | s = 0.3 | **3.62** | 674.941 | 186× | 609.633 | 168× |
|  | s = 0.4 | **3.741** | 664.363 | 177× | 635.302 | 169× |
|  | s = 0.5 | **3.691** | 662.802 | 179× | 603.838 | 163× |
|  | s = 0.6 | **3.619** | 659.336 | 182× | 611.441 | 168× |
|  | s = 0.7 | **3.596** | 648.885 | 180× | 689.137 | 191× |

Table 7: model1 varying $n_c$ and $n_d$ in high-dimensional setting

|  | Model | **DIFFEE** | FusedGLasso | Slower | Density Ratio | Slower |
|---|---|---|---|---|---|---|
| F1-score | $n_c = p/4, n_d = p/4$ | **0.008** | 0.008 | | 0 | |
|  | $n_c = p/4, n_d = p/2$ | **0.008** | 0.008 | | 0 | |
|  | $n_c = p/2, n_d = p/4$ | **0.016** | 0.008 | | 0 | |
|  | $n_c = p/2, n_d = p/2$ | **0.009** | 0.008 | | 0.009 | |
| Time (s) | $n_c = p/4, n_d = p/4$ | **3.647** | 696.742 | 191× | 398.226 | 109× |
|  | $n_c = p/4, n_d = p/2$ | **3.61** | 704.943 | 195× | 590.044 | 163× |
|  | $n_c = p/2, n_d = p/4$ | **3.609** | 697.858 | 193× | 408.149 | 113× |
|  | $n_c = p/2, n_d = p/2$ | **3.582** | 654.147 | 182× | 642.168 | 179× |

Table 8: model2 varying $n_c$ and $n_d$ in high-dimensional setting

|  | Model | DIFFEE | FusedGLasso | Slower | Density Ratio | Slower |
|---|---|---|---|---|---|---|
| F1-score | $n_c = p/4, n_d = p/4$ | **0.45** | 0.221 | | 0.065 | |
|  | $n_c = p/4, n_d = p/2$ | **0.45** | 0.226 | | 0.063 | |
|  | $n_c = p/2, n_d = p/4$ | **0.45** | 0.29 | | 0.065 | |
|  | $n_c = p/2, n_d = p/2$ | **0.45** | 0.203 | | 0.066 | |
| Time (s) | $n_c = p/4, n_d = p/4$ | **3.74** | 654.227 | 174× | 381.686 | 102× |
|  | $n_c = p/4, n_d = p/2$ | **3.748** | 654.822 | 174× | 484.77 | 129× |
|  | $n_c = p/2, n_d = p/4$ | **3.717** | 653.657 | 175× | 346.148 | 93× |
|  | $n_c = p/2, n_d = p/2$ | **3.528** | 657.147 | 186× | 494.066 | 140× |

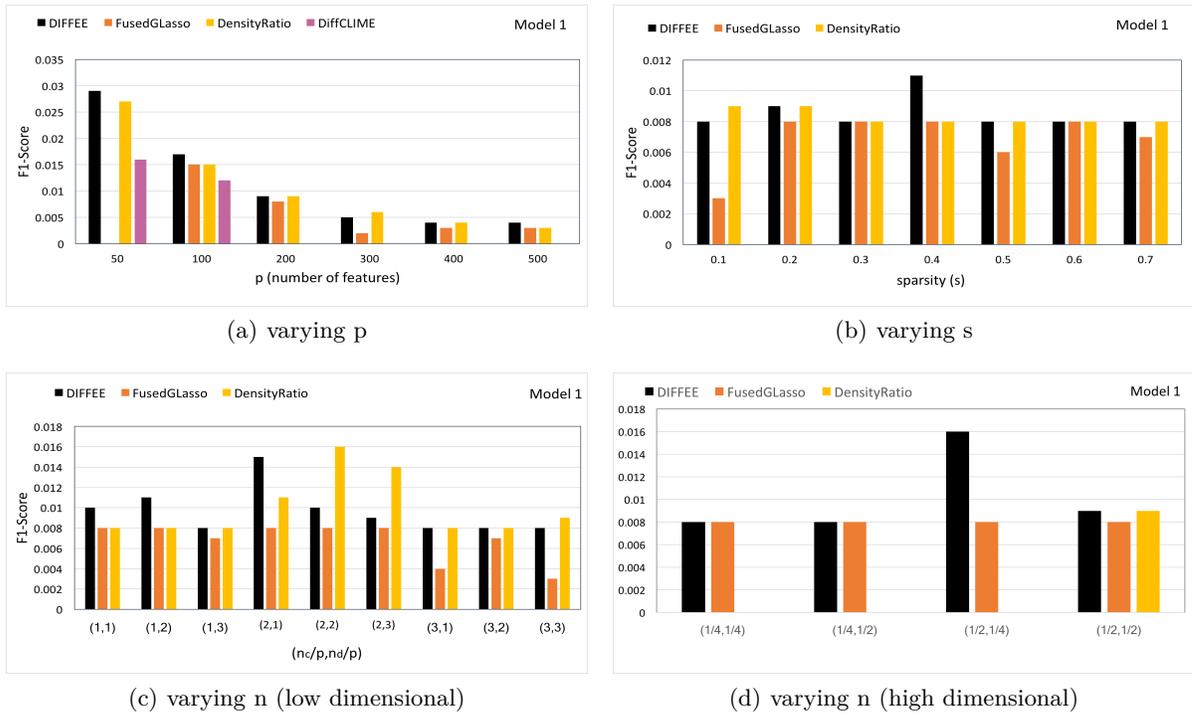

(a) varying p  (b) varying s  
(c) varying n (low dimensional)  (d) varying n (high dimensional)

Figure 5: F1-Score of DIFFEE and baseline methods for Simulated Model 1

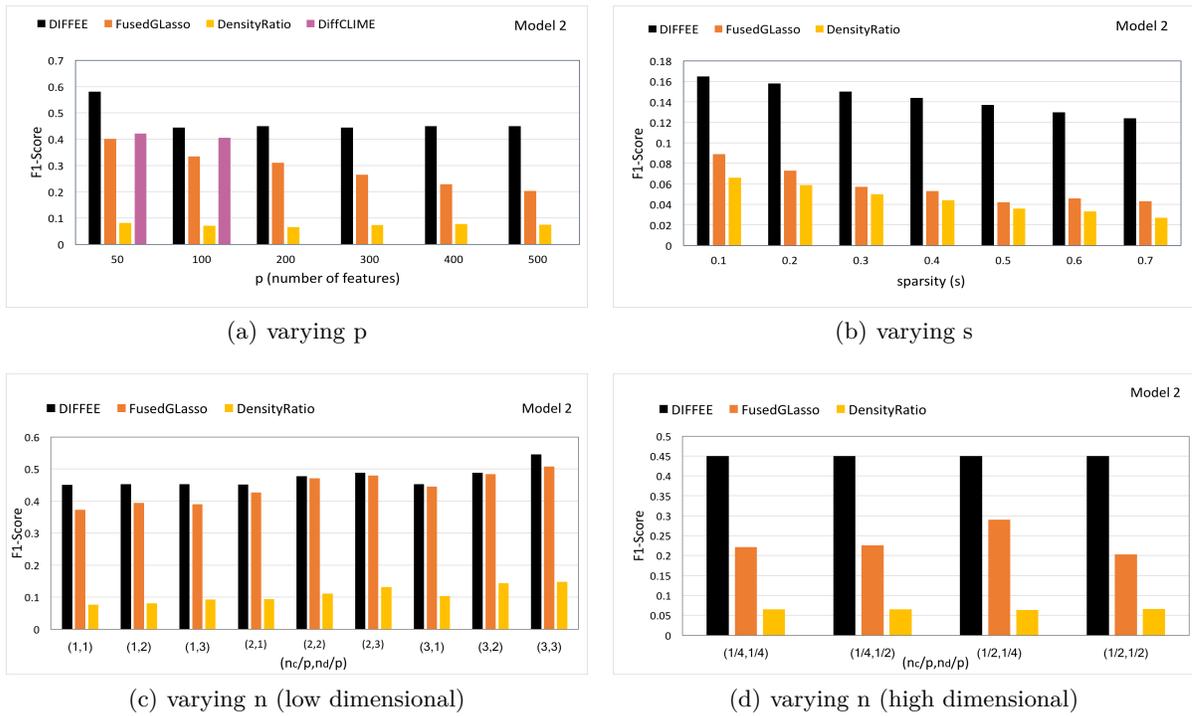

(a) varying p  (b) varying s  
(c) varying n (low dimensional)  (d) varying n (high dimensional)

Figure 6: F1-Score of DIFFEE and baseline methods for Simulated Model 2

Table 9: model1 varying $n_c$ and $n_d$ in low-dimensional setting

|  | Model | DIFFEE | FusedGLasso | Slower | Density Ratio | Slower |
|---|---|---|---|---|---|---|
| F1-score | $n_c = p, n_d = p$ | **0.01** | 0.008 | | 0.008 | |
| | $n_c = p, n_d = 2p$ | **0.011** | 0.008 | | 0.008 | |
| | $n_c = p, n_d = 3p$ | **0.008** | 0.007 | | 0.008 | |
| | $n_c = 2p, n_d = p$ | **0.015** | 0.008 | | 0.011 | |
| | $n_c = 2p, n_d = 2p$ | 0.01 | 0.008 | | **0.016** | |
| | $n_c = 2p, n_d = 3p$ | 0.009 | 0.008 | | **0.014** | |
| | $n_c = 3p, n_d = p$ | **0.008** | 0.004 | | 0.008 | |
| | $n_c = 3p, n_d = 2p$ | **0.008** | 0.007 | | 0.008 | |
| | $n_c = 3p, n_d = 3p$ | 0.008 | 0.003 | | **0.009** | |
| Time (s) | $n_c = p, n_d = p$ | **3.643** | 691.581 | 189× | 838.863 | 230× |
| | $n_c = p, n_d = 2p$ | **3.569** | 1023.507 | 286× | 1468.593 | 411× |
| | $n_c = p, n_d = 3p$ | **3.62** | 1319.354 | 364× | 2054.228 | 567× |
| | $n_c = 2p, n_d = p$ | **3.578** | 700.539 | 195× | 932.511 | 260× |
| | $n_c = 2p, n_d = 2p$ | **3.568** | 875.55 | 245× | 1291.795 | 362× |
| | $n_c = 2p, n_d = 3p$ | **3.553** | 1406.44 | 395× | 2224.744 | 626× |
| | $n_c = 3p, n_d = p$ | **3.587** | 696.087 | 194× | 882.885 | 246× |
| | $n_c = 3p, n_d = 2p$ | **3.578** | 725.195 | 202× | 1464.343 | 409× |
| | $n_c = 3p, n_d = 3p$ | **3.592** | 1264.346 | 351× | 2191.003 | 609× |

Table 10: model2 varying $n_c$ and $n_d$ in low-dimensional setting

|  | Model | DIFFEE | FusedGLasso | Slower | Density Ratio | Slower |
|---|---|---|---|---|---|---|
| F1-score | $n_c = p, n_d = p$ | **0.45** | 0.372 | | 0.076 | |
| | $n_c = p, n_d = 2p$ | **0.453** | 0.394 | | 0.081 | |
| | $n_c = p, n_d = 3p$ | **0.452** | 0.39 | | 0.092 | |
| | $n_c = 2p, n_d = p$ | **0.451** | 0.426 | | 0.093 | |
| | $n_c = 2p, n_d = 2p$ | **0.477** | 0.471 | | 0.111 | |
| | $n_c = 2p, n_d = 3p$ | **0.488** | 0.479 | | 0.131 | |
| | $n_c = 3p, n_d = p$ | **0.452** | 0.445 | | 0.103 | |
| | $n_c = 3p, n_d = 2p$ | **0.488** | 0.484 | | 0.143 | |
| | $n_c = 3p, n_d = 3p$ | **0.546** | 0.508 | | 0.148 | |
| Time (s) | $n_c = p, n_d = p$ | **3.658** | 707.735 | 193× | 714.371 | 195× |
| | $n_c = p, n_d = 2p$ | **3.746** | 688.608 | 183× | 1192.792 | 318× |
| | $n_c = p, n_d = 3p$ | **3.673** | 676.806 | 184× | 1707.516 | 464× |
| | $n_c = 2p, n_d = p$ | **3.69** | 673.112 | 182× | 723.656 | 196× |
| | $n_c = 2p, n_d = 2p$ | **3.691** | 676.597 | 183× | 1164.175 | 315× |
| | $n_c = 2p, n_d = 3p$ | **3.57** | 677.65 | 189× | 1830.678 | 512× |
| | $n_c = 3p, n_d = p$ | **3.692** | 673.364 | 182× | 717.752 | 194× |
| | $n_c = 3p, n_d = 2p$ | **3.692** | 682.499 | 184× | 1090.64 | 295× |
| | $n_c = 3p, n_d = 3p$ | **3.732** | 719.733 | 192× | 1739.274 | 466× |